\documentclass[runningheads]{llncs}
\usepackage{graphicx}
\usepackage[switch]{lineno}
\usepackage{caption} % DO NOT CHANGE THIS AND DO NOT ADD ANY OPTIONS TO IT
\usepackage{subfigure}
\usepackage{amsmath,amssymb,amsfonts}
\usepackage{textcomp}
\usepackage{xcolor}

\newcommand{\yy}[1]{\textcolor{black}{#1}}

\usepackage{multirow}
\usepackage{multicol}
\usepackage[ruled,vlined,linesnumbered,boxed]{algorithm2e}
\usepackage{algpseudocode}
\usepackage{booktabs}
\begin{document}
\title{Unified Robust Training for Graph Neural Networks against Label Noise}
%\orcidID{0000-1111-2222-3333}
\author{}
\institute{}
\author{Yayong Li\inst{1} \and
Jie Yin\inst{2}\and
Ling Chen\inst{1}}
% %
% \authorrunning{Li et al.}
\institute{Faculty of Engineering and Information Technology, University of Technology Sydney, Australia\\ \email{Yayong.Li@student.uts.edu.au, Ling.Chen@uts.edu.au} \and
Discipline of Business Analytics, The University of Sydney, Australia\\
\email{jie.yin@sydney.edu.au}}
\maketitle              % typeset the header of the contribution
\begin{abstract}
Graph neural networks (GNNs) have achieved state-of-the-art performance for node classification on graphs. The vast majority of existing works assume that genuine node labels are always provided for training. However, there has been very little research effort on how to improve the robustness of GNNs in the presence of label noise. Learning with label noise has been primarily studied in the context of image classification, but these techniques cannot be directly applied to graph-structured data, due to two major challenges---\textit{label sparsity} and \textit{label dependency}---faced by learning on graphs. In this paper, we propose a new framework, UnionNET, for learning with noisy labels on graphs under a semi-supervised setting. Our approach provides a unified solution for robustly training GNNs and performing label correction simultaneously. The key idea is to perform label aggregation to estimate node-level class probability distributions, which are used to guide sample reweighting and label correction. Compared with existing works, UnionNET has two appealing advantages. First, it requires no extra clean supervision, or explicit estimation of the noise transition matrix. Second, a unified learning framework is proposed to robustly train GNNs in an end-to-end manner. Experimental results show that our proposed approach: (1) is effective in improving model robustness against different types and levels of label noise; (2) yields significant improvements over state-of-the-art baselines.

\end{abstract}
\section{Introduction}
Nowadays, graph-structured data is being generated across many high-impact applications, ranging from financial fraud detection in transaction networks to gene interaction analysis, from cyber security in computer networks to social network analysis. To ingest rich information on graph data, it is of paramount importance to learn effective node representations that encode both node attributes and graph topology. To this end, graph neural networks (GNNs) have been proposed, built upon the success of deep neural networks (DNNs) on grid-structured data (e.g., images, etc.). GNNs have abilities to integrate both node attributes and graph topology by recursively aggregating node features across the graph. GNNs have achieved state-of-the-art performance on many graph related tasks, such as node classification or link prediction. 

%to consider the underlying graph as a computation graph and 

The core of GNNs is to learn neural network primitives that generate node representations by passing, transforming, and aggregating node features from local neighborhoods~\cite{gilmer2017neural}. As such, nearby nodes would have similar node representations~\cite{velivckovic2017graph}. By generalizing convolutional neural networks to graph data, graph convolutional networks (GCNs)~\cite{kipf2017semi} define the convolution operation via a neighborhood aggregation function in the Fourier domain. The convolution of GCNs is a special form of Laplacian smoothing on graphs~\cite{li2018deeper}, which mixes the features of a node and its nearby neighbors. However, this smoothing operation can be disrupted when the training data is corrupted with label noise. As the training proceeds, GCNs would completely fit noisy labels, resulting in degraded performance and poor generalization. Hence, one key challenge is how to improve the robustness of GNNs against label noise. 

%Deep neural networks (DNNs) have recently achieved state-of-the-art performance on many classification tasks. Their remarkable performance is largely attributed to the collection of large datasets with human annotated labels. However, it is extremely time-consuming and costly to annotate a large quantity of data with high-quality labels. This is particularly the case for graph-structured data, where samples are not independent of each other, but form a graph structure with edges representing the relationships between samples.

%anchor node and its neighbors
%In the sense of \cite{li2018deeper}, graph convolutional networks is actually a special Laplacian smoothing, which mixes the features of the adjacent nodes. This mixing operation allow the nodes in the same cluster have a similar representations, easing the follow-up classification tasks. 
 % The significance of label noise is expected to increase as larger datasets are prepared for training DNNs. 
Learning with noisy labels has been extensively studied on image classification. Label noise naturally stems from inter-observer variability, human annotator's error, and errors in crowdsourced annotations~\cite{karimi2020deep}. Existing methods attempt to correct the loss function by directly estimating a noise transition matrix~\cite{patrini2017making,vahdat2017toward}, or by adding extra layers to model the noise transition matrix~\cite{sukhbaatar2014training,goldberger2016training}. However, it is difficult to accurately estimate the noise transition matrix particularly with a large number of classes. 
Alternative methods such as MentorNet~\cite{jiang2017mentornet} and Co-teaching~\cite{han2018co} seek to separate clean samples from noisy samples, and use only the most likely clean samples to update model training. %MentorNet~\cite{jiang2017mentornet} and Co-teaching~\cite{han2018co} are two such representative methods. 
%MentorNet~\cite{jiang2017mentornet} pre-trains a teacher network to select clean samples to guide network training, which requires the access to a set of clean samples. Co-teaching~\cite{han2018co} trains two peer networks to select small-loss samples within each mini-batch to train each other. %The threshold choice for selecting small-loss samples yet heavily depends on the estimated noise rate. 
Other methods~\cite{arazo2019unsupervised,ren2018learning} reweight each sample in the gradient update of the loss function, according to model's predicted probabilities. However, they require a large number of labeled samples or an extra clean set for training. Otherwise, reweighting %based on model predictions 
would be unreliable and result in poor performance.

%Unlike the aforementioned methods that study the problem of label noise under a fully supervised setting, we focus our research on the semi-supervised learning, where only a small fraction of nodes have noisy labels while the rest of nodes are unlabeled. %accompanying with a large number of unlabeled nodes, e.g. only 3.6\% of node are labeled in Citeseer dataset. 
%In this situation, a robust classifier is tougher to attain due to two factors: 1) The classifier is rather sensitive to the number of the labeled node, and the rapid reduction of labeled nodes will result in rapidly declined classification accuracy. Thus, the existing label information become the very precious resource that should be fully exploited rather than simply excluding the suspect labels. Any methods that do not take this into consideration will degrade the performance, or even fail to obtain acceptable results. 2) How to use the unlabeled node to prevent label errors propagation. 

The aforementioned learning techniques, however, cannot be directly applied to tackle label noise on graphs. This is attributed to two significant challenges. (1) \textbf{Label sparsity}: graphs with inter-connected nodes are arguably harder to label than individual images. Very often, graphs are sparsely labeled, with only a small set of labeled nodes provided for training. Hence, we cannot simply drop ``bad nodes'' with corrupted labels like previous methods using ``small-loss trick''~\cite{han2018co,jiang2017mentornet}. (2) \textbf{Label dependency}: graph nodes exhibit strong label dependency, so nodes with high structural proximity (directly or indirectly connected) tend to have a similar label. This presses a strong need to fully exploit graph topology and sparse node labels when training a robust model against label noise.

To tackle these challenges, we propose a novel approach for robustly learning GNN models against noisy labels under semi-supervised settings. Our approach provides a \underline{uni}fied r\underline{o}bust trai\underline{n}ing framework for graph neural \underline{net}works (UnionNET) that performs sample reweighting and label correction simulatenously. The core idea is twofold: (1) leverage random walks to perform label aggregation among nodes with structural proximity. (2) estimate node-level class distribution to guide sample reweighting and label correction. %\yy{Intuitively, when labeled nodes are scarce, one node may have very few or no labeled neighbors. %making it difficult to examine the reliability of its label. 
%Thus, we use random walks to expand a node's structural context, and aggregate labels from this context to infer node-level class distribution more accurately.} 
Intuitively, noisy labels could cause disordered predictions around context nodes, thus its derived node class distribution could in turn reflect the reliability of given labels. This provides an effective way to assess the reliability of given labels, guided by which sample reweighting and label correction are expected to weaken unreliable supervision and encourage label smoothing around context nodes. We verify the effectiveness of our proposed approach through experiments and ablation studies on real-world networks, demonstrating its superiority over competitive baselines.

%We expect that the reweighting mechanism guided by context nodes could encourage label smoothing and weaken unreliable supervisions.

%UnionNET has two appealing advantages. First, unlike most of previous work, it does not require extra clean supervision, or explicit estimation of the noise transition matrix. Second, a unified framework is proposed to train robust GNNs in an end-to-end manner. UnionNET is flexible to be instantiated with off-the-shelf semi-supervised GNNs to improve the robustness against label noise. Our proposed method is shown to be effective in improving model robustness against label noise with different types and noise rates. %outperforming state-of-the-art by a large margin on node classification.

% The main contributions of this work are threefold:
% \begin{itemize}
%     \item \textcolor{red}{We are the first to study the problem of learning with noisy labels on graphs under a semi-supervised setting.} 
%     \item We propose a unified learning framework for robustly training GNNs and correcting noisy labels simultaneously.
%     \item Experiments and ablation studies verify the effectiveness of our method and its superiority over competitive baselines.
% \end{itemize}

\section{Related Work}
%This section reviews related work in the space of learning with noisy labels and graph neural networks. 

\subsection{Learning with Noisy Labels}
Learning with noisy labels has been widely studied in the context of image classification. The first line of research focuses on correcting the loss function, by directly estimating the noise transition matrix between noisy labels and ground true labels~\cite{patrini2017making,vahdat2017toward}, or adding an extra softmax layer to estimate the noise transition matrix~\cite{sukhbaatar2014training,goldberger2016training}.
%The correction can be categorized in two types. The first type focuses on estimating the noise transition matrix between noisy labels and ground true labels. 
%Patrini et al.~\cite{patrini2017making} introduced the noise transition matrix to the loss function, and proposed a two-level estimation approach when the matrix is not known a priori. Other methods~\cite{xiao2015learning,vahdat2017toward} used a graphical model to capture the relationship between noise labels and ground-truth labels, An EM-like algorithm was proposed to infer the true labels. Both methods yet require the access to a set of clean samples. 
%The second type proposed to add an extra softmax layer to estimate the probability of the ground-truth label flapping to noisy labels~\cite{sukhbaatar2014training,goldberger2016training}. 
However, it is non-trivial to estimate the noise transition matrix accurately. %, in particular, with a large number of classes. %Considering the fact that the standard cross entropy loss implicitly put more emphasis on difficult samples and the mean absolute error would significantly slow down the convergence of model training, 
\cite{zhang2018generalized} used the negative Box-Cox transformation to improve the robustness of standard cross entropy loss but with worse converging capacity.
%This loss function could be seen as a generalization of the two loss functions, and can be easily adapted to other networks.
%Patrini et al.~\cite{patrini2017making} proposed a forward and a backward loss correction method by introduce a transition matrix to the loss functions. This method is subject to a known noise transition matrix. If not, it has to be estimated, which is not reliable when there exists a large number of classes.
%Arash et al.~\cite{vahdat2017toward} proposed a graphical model to capture the relationship between noise labels and ground-truth labels, by modeling ground-truth labels as latent variables and noise labels as observed variables. Then an EM-like method was used to infer the true labels. The similar EM-like method was also seen in~\cite{xiao2015learning}. Ren et al. \cite{ren2018learning} employed a meta-learning algorithm which allows the network to put more emphasis on the samples that have the closest gradient directions with the clean data. However, all the aforementioned methods require the access to a set of clean samples for extra supervision.
%\cite{sukhbaatar2014training,goldberger2016training} added an extra layer to estimate the probability of the ground-truth label flapping to noisy labels. However, it is usually difficult to obtain a precise transition matrix, and moreover, these methods implicitly assumed that the transition probability is independent of individual samples.
The second line of approaches seek to separate clean samples from noisy ones, and use only the most likely clean samples to guide network training.
%Some approaches proposed to train their networks using the selected samples that are most lkely to be clean. 
MentorNet~\cite{jiang2017mentornet} pre-trains an extra network on a clean set to select clean samples. % to update the network training.  %based on their training losses. 
%When the clean set was unavailable, MentorNet had to rely on a predifined curriculum, % (e.g., self-paced curriculum)
%yielding suboptimal results.
%\cite{guo2018curriculumnet} learnt a distance density-based curriculum to prioritize low-complexity samples. 
%The curriculum, however, is designed using features extracted from a pre-trained network trained on all noisy samples. Thus, the results become unreliable when noise rates are high. 
Co-teaching~\cite{han2018co} trains two peer networks to select small-loss samples %within each mini-batch 
to train each other. %It is further improved by~\cite{yu2019does} that updates the network only with small-loss samples that two networks disagree with. Both methods, 
%However, choosing the threshold for selecting samples relies on the estimated noise rate. 
Decoupling~\cite{malach2017decoupling} updates two networks using only samples with which they disagree. In our setting with very few labeled nodes, we cannot simply drop ``bad nodes" as they are still useful to infer the labels of nearby nodes.
%-- find the samples that two networks most disagree.
%Malach et al. \cite{malach2017decoupling} proposed a \textit{decoupling} method which maintained two networks but only updated their parameters using the samples with which the two networks disagreed.
%MentorNet~\cite{jiang2017mentornet} proposed a data-driven curriculum learning network, which pre-trained an LSTM network to select the most probably correct samples to guide the network training based on their training losses. However if clean data is unavailable, it would have to resort to a predefined curriculum to train the LSTM network, which would only obtain a suboptimal results.
%Han et al.\cite{han2018co} proposed a co-teaching framework based on the assumption that two networks with different initialized parameters could filter different kind of noise. It trained two networks using the samples which has a small loss in the peer network. But the threshold of small loss heavily rely on the estimated noise rate.  This method was then improved by Yu et al. \cite{yu2019does}, which only used the samples with small loss and different predictions to teach the peer network. However, although the selected training samples is less noisy, they reduce the number if the training samples and discard useful information of given labels, which would degrade the performance especially when the labeled data is not enough.
The third category takes a reweighting approach. %based on model's predicted probabilities. 
\cite{arazo2019unsupervised} utilized a two-component Beta Mixture Model to estimate the probability of a sample being mislabeled, which is used to reweight the sample in the gradient update. It was further improved by combining with~\textit{mixup augmentation}~\cite{zhang2017mixup}. \cite{ren2018learning} proposed a meta-learning algorithm that allowed the network to put more weights on the samples with the closest gradient directions with the clean data. Unlike these reweighting methods that rely on the predicted probabilities, our method assigns weights to each node by leveraging topology structure, which is less prone to label noise.
%assigns different weights to each node during backpropagation. However, our method is essentially distinct from previous work in that our method calculates the weight of each node based on topology structure in its neighborhood, while previous work relies on the predicted probabilities by the trained model itself. 
%However, a model that is created totally based on the training loss is not sufficient enough to determine whether the label is correct. Moreover, as the model put different weights on the samples in the process of gradient updating, samples with larger weight would have smaller loss while those with small weights would have larger loss, thus the accuracy of this model would degrade. 
%After that, NT et al.\cite{nt2019learning} transferred this method into graph neural networks to deal with the noise problem in graph classification tasks. 
Several other methods are concerned with the problem of label correction. \cite{han2019deep} chose class prototypes based on sample distance density to correct labels, incurring significant computational overhead. \cite{tanaka2018joint} proposed a self-training approach to correct the labels. %by alternatively updating network parameters and labeling towards the gradient descending direction of the training loss. 
However, 
%this method is completely dependent on predicted labels, 
this method discards the original given labels, leading to degraded performance with high noise rates. Our work integrates sample reweighting with label correction, yielding remarkable gains with high noise rates.  

%a unified solution for integrating robust training with label correction, 

%Similarly, Yi et al.~\cite{yi2019probabilistic} modeled the true labels as a probability distribution and update it together with network parameters. This method faces the same problem with Tanaka et al. ~\cite{tanaka2018joint}, and moreover, it needs to fine-tune its hyper-parameters for different noise rates. 

\subsection{Graph Neural Networks}

GNNs have emerged as a new class of deep learning models on graphs. %~\cite{kipf2017semi,velivckovic2017graph}. 
%The principle of GNNs is to learn node representations by recursively aggregating and compressing the continuous feature vectors from local neighborhoods. The generated node representations can then be used as input to any differentiable prediction layer, for example, a softmax layer for node classification. %The whole model can be trained in an end-to-end architecture. 
Various types of GNNs, such as GCN~\cite{kipf2017semi}, graph attention network (GAT)~\cite{velivckovic2017graph}, GraphSAGE~\cite{hamilton2017inductive}, are proposed in recent years. %They differ mainly in the way how features are aggregated in local neighborhood. 
These models have shown competitive results on node classification, assuming that genuine node labels are provided for training purposes. 
%Thus, it shares the same limitation with similar methods on image data that the noise transition matrix is not easy to be accurately estimated. 
%few shot on image data
%Iscen et al.~\cite{Iscen2019GraphCN} propose a method that builds a GCN for each class and that uses GCN's inferred probability as a relevance measure to select clean samples. However, it requires a set of clean data to provide additional supervision.
%Following~\cite{kipf2017semi}, 
To date, there has been little research work on robustly training a GNN against label noise. \cite{nt2019learning} studied the problem of learning GNNs with symmetric label noise. This method adopts a backward loss correction~\cite{patrini2017making} for graph classification. \cite{de2020analysis} analyzed the robustness of traditional collective node classification methods on graphs (such as label propagation) towards random label noise, but it did not propose new solutions to tackle this problem. %In this work, we focus on semi-supervised node classification with both symmetric and asymmetric label noise. %Given a network with a small set of nodes being labeled and the rest remaining unlabeled, we aim to learn a robust model that effectively predict the class labels for the unlabeled nodes. 
To the best of our knowledge, our work is the first to study the problem of learning robust GNNs for semi-supervised node classification on graphs with both symmetric and asymmetric label noise. Our method provides a unified learning framework and does not require explicit estimation of the noise transition matrix.

\section{Problem Definition}
Given an undirected graph $G=\{\mathcal{V}, \mathcal{E}, \mathbf{X}\}$, where $\mathcal{V}$ denotes a set of $n$ nodes, and $\mathcal{E}$ denotes a set of edges connecting nodes. $\mathbf{X} = [\mathbf{x}_1, \mathbf{x}_2, \ldots, \mathbf{x}_n]^T\in \mathcal{R}^{n\times d}$ denotes the node feature matrix, where $\mathbf{x}_i \in \mathcal{R}^d$ is $d$-dimensional feature vector of node $v_i$. Let $A \in \mathcal{R}^{n \times n}$ denote the adjacent matrix. % which is nonnegative. 

We consider semi-supervised node classification, where only a small fraction of nodes are labeled. Let $\mathcal{L}=\{(\mathbf{x}_i, \mathbf{y}_i)\}_{i=1}^{|L|}$ denote the set of labeled nodes, where $\mathbf{x}_i$ is feature vector of node $v_i$, and $\mathbf{y}_i=\{y_{i1},y_{i2},\dots,y_{im}\}$ is the one-hot encoding of node $v_i$'s class label, with $y_{ij} \in \{0, 1\}$ and $m$ being the number of classes. The rest of nodes belong to the unlabeled set $\mathcal{U}$. Under the GNN learning framework, the aim is to learn a representation $\mathbf{h}_{\mathbf{x}_i}$ for each node $v_i$ such that its class label can be correctly predicted by $f(\mathbf{h}_{\mathbf{x}_i})$. For node classification, the standard cross entropy loss is used as the objective function:
\begin{equation}\begin{small}
\label{eq:standardCE}
    \mathcal{J}(f(\mathbf{h}_{\mathbf{x}}), \mathbf{y}) = - \sum_{i\in |L|} \sum_{j\in m} \mathbf{y}_{ij} \log(f(\mathbf{h}_{\mathbf{x}_i})_j).
    \end{small}
\end{equation}

However, when class labels in $\mathcal{L}$ are corrupted with label noise, the standard cross entropy would cause the GNN training to overfit incorrect labels, and in turn lead to degraded classification performance. Therefore, in our work, we aim to train a robust GNN model that is less sensitive to label noise. 

Formally, given a small set of noisy labeled nodes $\mathcal{L}$, we aim to: (1) learn node representations $\mathbf{h}$ for all nodes $\mathcal{V}$, and (2) learn a model $f(\mathbf{h})$ to predict the labels of unlabeled nodes in $\mathcal{U}$ with maximum classification performance.

\section{The UnionNET Learning Framework}
%This section begins with an overview of our proposed framework and discusses its key components and loss function.

\begin{figure*}
\centering
  \includegraphics[width=\textwidth]{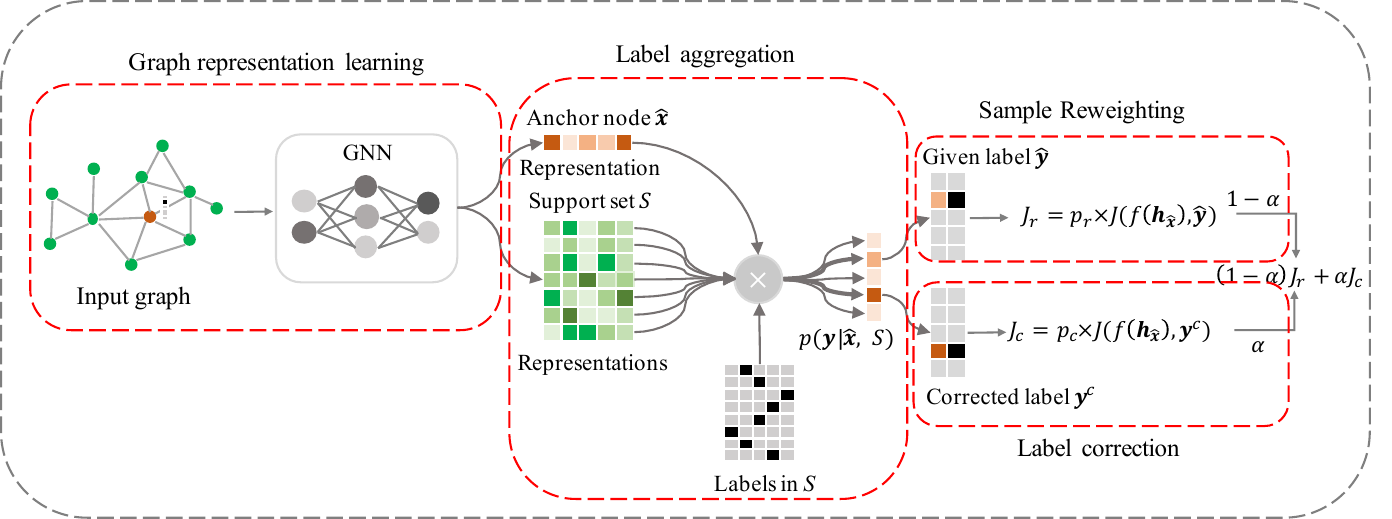}
  \caption{Overview of the UnionNET Framework. The key idea is to infer the reliability of the given labels through estimating node-level class probability distributions via label aggregation. Based on this, the corresponding label weights and corrected labels are obtained to update model parameters during training.}
  \label{fig:framework}
\end{figure*}

%\subsection{Framework Overview}
\yy{To effectively tackle label noise on graphs, one desirable solution should consider the following key aspects. First, since only a small set of labeled nodes are available for training, we cannot simply drop ``bad nodes'' using ``small-loss trick''~\cite{han2018co,jiang2017mentornet}. Second, graph nodes that share similar structural context exhibit label dependency. Thus, we propose a unified framework, UnionNET, for robustly training a GNN and performing label correction, as shown in Fig~\ref{fig:framework}.}

%Even if one node's label is corrupted, its corrupted label can still be informative to estimate the labels of nearby nodes in the neighborhood. 
%UnionNET mainly consists of four components, i.e., graph representation learning, label aggregation, sample reweighting, and label correction. We visualize the main procedures in Fig.~\ref{fig:framework}. 

%GCN~\cite{kipf2017semi} is first applied to generate node representation and predicted label for each node. Note that here, GCN can be replaced by other semi-supervised GNN models.

Taking a given graph as input, a GNN is first applied to learn node representations and generate the predicted label for each node. Then, label information is aggregated to estimate a class probability distribution per node. This aggregation is operated on a support set constructed by collecting context nodes with high structural proximity. According to node-level class probability distributions, our algorithm generates label weights and corrected labels for each labeled node. Those corrected labels generated from the support set could potentially provide extra ``correct'' supervision. Taken together, both given labels reweighted by label weights and corrected labels are used to update model parameters. % during model training.

\subsection{Label Aggregation}
%due to the difficulty and expensive cost of labeling graph nodes, our proposed framework

%in the process of label prediction. %Based on this, graph neural networks push similar nodes together while drive dissimilar nodes away from each other in the low-dimensional embedding space.

%collects all the label information together from the most associative nodes, and then aggregate them to produce a class probability distribution. This class probability distribution would be used to guide label prediction in the follow-up tasks.

On graphs, it is well studied that nodes with high structural proximity %, e.g. the nodes with direct connections or sharing the same neighbors, 
tend to have the same labels~\cite{lu2003link,zhu2003semi}. The supervision from noisy labels however disrupt such label smoothness around context nodes. Nevertheless, their smoothness degree could provide a reference to assess the reliability of given labels. Hence, we design a label aggregator that aggregates label information for each labeled node from its context nodes to estimate its class probability distribution. Specifically, we perform random walks to collect context nodes with higher-order proximity. For each \yy{labeled} node $\hat{\mathbf{x}} \in \mathcal{L}$, called \textit{anchor node}, we construct a \textit{support set} of size $k$, denoted as $S=\{(\mathbf{x}_i, \mathbf{y}_i)|\hat{\mathbf{x}}\}^k$, where $\mathbf{x}_i$ is the supportive node in $S$ and $\mathbf{y}_i$ is one-hot encoding of $\mathbf{x}_i$'s class label. %Inspired by DeepWalk~\cite{perozzi2014deepwalk}, 
%A stream of random walks, by visiting different parts of the network, could collect the most associated nodes together. With regard to their labels, 
%\yy{Those nodes sharing more neighbors would appear more times than others in $S$.} 
During a random walk, if node $\mathbf{x}_i \in \mathcal{L}$, the given label $\mathbf{y}_i$ is collected in $S$. Otherwise, the predicted label is used. % as $\mathbf{y}_i$.

%from GCN is used as $y_i$. 

%and thus provide a credible reference for the anchor node to assess the reliability of given label. 

% = C_S(\hat{\mathbf{x}})$ 
Given anchor node $\hat{\mathbf{x}}$ and its support set $\mathcal{S}$, we derive a node-level class probability distribution $P(\mathbf{y}|\hat{\mathbf{x}}, S)$ over $m$ classes. It signifies the probabilities of the anchor node belonging to $m$ classes in reference of its support set. Particularly, we specify a non-parametric attention mechanism given by, %which is formalized in its simplest form:  %Eq.(\ref{attention}):
\begin{equation}
\label{full_P}
    P(\mathbf{y}|\hat{\mathbf{x}}, S) = \sum_{\mathbf{x}_i \in S} \mathcal{A}(\hat{\mathbf{x}}, \mathbf{x}_i)\mathbf{y}_i
    =\sum_{\mathbf{x}_i \in S} \frac{\exp(\mathbf{h}_{\mathbf{x}_i}^T \mathbf{h}_{\hat{\mathbf{x}}})}{\sum_{\mathbf{x}_j \in S} \exp(\mathbf{h}_{\mathbf{x}_j}^T \mathbf{h}_{\hat{\mathbf{x}}})}\mathbf{y}_i.
\end{equation}
Here, the probability of the anchor node belonging to each class is calculated according to its proximity with nearby nodes in the support set. %$A(\hat{x}, S)$ can also be considered as a density estimator, which approaches zero for the furthest nodes $x_i$ from the $\hat{x}$ according to some distance metric and an appropriate constant otherwise. In that case, $C_S(\hat{x})$ could retrieve the most reliable labels from the associative set.% 
We define the proximity as the inner product in the embedding space, and apply softmax to measure the contribution made by each label in the support set to estimating the anchor node' class probability distribution. 
%Thus, Eq.(\ref{attention}) is rewritten as: %the following form:
% \begin{equation}
% \label{full_P}
%     P(\mathbf{y}|\hat{\mathbf{x}}, S) = \sum_{\mathbf{x_i} \in S} \frac{\exp(\mathbf{h}_{\mathbf{x}_i}^T \mathbf{h}_{\hat{\mathbf{x}}})}{\sum_{\mathbf{x}_j \in S} \exp(\mathbf{h}_{\mathbf{x}_j}^T \mathbf{h}_{\hat{\mathbf{x}}})}\mathbf{\mathbf{y}_i},
% \end{equation}
%where $\mathbf{h}_{\mathbf{x}}$ denotes node $\mathbf{x}$'s representation. 
In the support set, if a node has a higher similarity with the anchor node (i.e., higher inner product), its label would contribute more to $P(\mathbf{y}|\hat{\mathbf{x}}, S)$, and vice verse. 
%Finally, the labels of the most similar nodes in the support set would dominate the probability distribution. 
%Eq.(\ref{full_P}) could also be regarded as a weighted vote mechanism.
This simple yet effective mechanism estimates a class probability distribution for each node, which is used to guide sample reweighting and label correction. 

%Given the set S and anchor node $\hat{x}$, this simple mechanism is able to generate a sufficiently discriminative probability distribution.  

\subsection{Sample Reweighting}
\label{reweight}

For GNNs, the standard cross entropy loss implicitly puts more emphasis on the samples for which the predicted labels disagree with the provided labels during gradient update. This mechanism enables faster convergence and better fitting to the training data. However, if there exist corrupted labels in the training set, this implicit weighting scheme would conversely push the model to overfit noisy labels, leading to degraded performance~\cite{zhang2018generalized}. To mitigate this, we devise a reweighting scheme for each node according to the reliability of its given label, so that the loss of reliable labels could contribute more during gradient update.
%\textbf{and the loss of corrupted ones could contribute less }
%could back propagate more while that of suspect ones back propagate less in the gradient update. 

Specifically, we define the reweighting score of anchor node $\hat{\mathbf{x}}$ as:
\begin{equation}
\label{reweight}
    p_r(\hat{\mathbf{y}}|\hat{\mathbf{x}}, S) = \sum_{\mathbf{x}_i \in S, \mathbf{y}_i = \hat{\mathbf{y}}} \frac{\exp(\mathbf{h}_{\mathbf{x}_i}^T \mathbf{h}_{\hat{\mathbf{x}}})}{\sum_{\mathbf{x}_j \in S} \exp(\mathbf{h}_{\mathbf{x}_j}^T \mathbf{h}_{\hat{\mathbf{x}}})}\mathbf{y}_i.
\end{equation}
The loss function for the labeled nodes is thus defined as:

\begin{equation}
\label{loss_r}
    \mathcal{J}_{r} = - \sum_{\hat{\mathbf{x}} \in \mathcal{L}} p_r(\hat{\mathbf{y}}|\hat{\mathbf{x}},S)\times \hat{\mathbf{y}}\log(f(\mathbf{h}_{\hat{\mathbf{x}}})),
\end{equation}
    %\sum_{\hat{\mathbf{x}} \in \mathcal{L}} p_r(\hat{\mathbf{y}}|\hat{\mathbf{x}},S)\times \hat{\mathbf{y}}log(f(\mathbf{h}_{\hat{\mathbf{x}}}))
%This reweighting method takes effects based on the aggregated label information, which strengthen 
where $p_r(\hat{\mathbf{y}}|\hat{\mathbf{x}},S)$ is the weight imposed on each labeled node $\hat{\mathbf{x}}$ according to the aggregated label information. If the given label $\hat{\mathbf{y}}$ is highly consistent with nearby labels, its gradient would be back-propagated as it is. Otherwise, it would be penalized by the weight during back-propagation.

\subsection{Label Correction}
The reweighting method reduces the sensitivity of the standard cross entropy to noisy labels, and boosts the robustness of the model. As labeled nodes are limited for training, we also augment the set of labeled nodes by correcting noisy labels. Accordingly, we define the label correction loss as
%is defined as Eq.(\ref{loss_c}) and Eq.(\ref{correction}):
%However, this also reduce the accessible label information, which would degrade the performance in the case of scarce labels. 
%We further utilize the support set to make label correction. 
\begin{equation}
\label{loss_c}
    \mathcal{J}_{c} = - \sum_{\hat{\mathbf{x}} \in \mathcal{L}} p_c(\mathbf{y}^c|\hat{\mathbf{x}},S)\times \mathbf{y}^c \log(f(\mathbf{h}_{\hat{\mathbf{x}}})),
\end{equation}
%    \sum_{\hat{\mathbf{x}} \in \mathcal{L}} p_c(\mathbf{y}^c|\hat{\mathbf{x}},S)\times \mathbf{y}^c log(f(g(\hat{\mathbf{x}})))
\begin{equation}
\label{correction}
    p_c(\mathbf{y}^c|\hat{\mathbf{x}}, S) = \max_{\mathbf{y}_i}P(\mathbf{y}_i|\hat{\mathbf{x}}, S) =  \max_{\mathbf{y}_i} \sum_{\mathbf{x}_i \in S} \frac{\exp(\mathbf{h}_{\mathbf{x}_i}^T \mathbf{h}_{\hat{\mathbf{x}}})}{\sum_{\mathbf{x}_j \in S} \exp(\mathbf{h}_{\mathbf{x}_j}^T \mathbf{h}_{\hat{\mathbf{x}}})} \mathbf{y}_i.
\end{equation}
This provides additional supervision for $\hat{\mathbf{x}}$ with the corrected label $\mathbf{y}^c$, encouraging it to have the same label with the most consistent one in its support set. %We can also view this reweighting and loss correction method under the guidance of support set as a label smoothing method.
This approach aggregates labels from context nodes via a linear combination based on their similarity in the embedding space. %Referring to these neighboring label information, this method 
It thus helps diminish the gradient update of corrupted labels, and boosts the supervision from consistent labels. %As such, the nodes with structural proximity are encouraged to have the same label, which alleviate the impact of incorrect labels. 

%The two reweighted loss functions are integrated in a linear manner:
%\begin{equation}
%    L_{sup} = (1 - \alpha) L_r + \alpha L_c,
%\end{equation}
%where $\alpha$ controls the trade-off between the two terms in the gradient update. 

%\subsubsection{Entropy loss} 
%The entropy loss is used to regularize label correction. When the corrected label is different from the given label, the network would receive supervision from two different labels, causing a bimodality in the predicted class distribution. 
%Thus, we regularize the entropy of the predicted labels by minimizing Eq.(\ref{entropy}), aiming to encourage to generate predictions with low uncertainty. %the estimated class distribution towards the most probable class and, The entropy loss is as follows:
%\begin{equation}
%\label{entropy}
%    L_e = - \sum_{i=1}^n \sum_{j=1}^m f(\mathbf{h}_{\mathbf{x}_{i}})_j \log ~f(\mathbf{h}_{\mathbf{x}_{i}})_j
%\end{equation}
   % \sum_{i=1}^N \sum_{j=1}^M g(\mathbf{x}_i)_j log ~g(\mathbf{x}_i)_j

%\subsubsection{Prior distribution loss} 
 
%In our semi-supervised setting where the labeled nodes of each class are often scarce, label noise would make the corrupted class distribution deviate from its original prior distribution over all training data. In some extreme cases, it might cause severe class imbalance problems, leading to a high prediction error. 
%In our proposed solution, intuitively, we operate label correction based on the estimated reliability of the given label based on the smoothness of predicted labels in the context nodes. 
However, in the presence of extreme label noise, this approach would produce biased correction that deviates far away from its original prior distribution over the training data. This bias could exacerbate the overfitting problem caused by noisy labels. To overcome this, we employ a KL-divergence loss between the prior and predicted distributions to push them as close as possible~\cite{tanaka2018joint}. It is given by:
\begin{equation}
\label{loss_p}
    \mathcal{J}_p = \sum_{j=1}^m p_j \log\frac{p_j}{\overline{f(\mathbf{h}_\mathbf{X})}_j},
\end{equation}
% \begin{equation}
% \label{prior}
%     \overline{f(\mathbf{h}_X)}_j = \frac{1}{|L|}\sum_{\mathbf{x}\in L}f(\mathbf{h}_{\mathbf{x}})
% \end{equation}
%\sum_{j=1}^M p_j log\frac{p_j}{\overline{f(g(X))}_j}
Where $p_j$ is the prior probability of class $j$ in $\mathcal{L}$, and $\overline{f(\mathbf{h}_\mathbf{X})}_j=\frac{1}{|L|}\sum_{\mathbf{x}\in \mathcal{L}}f(\mathbf{h}_{\mathbf{x}})_j$ is the mean value of predicted probability distribution on the training set. %, which is calculated as:

%\textcolor{red}{(i.e. $1/m$ in our situation)}

\subsection{Model Training}
The training of UnionNET is given in Algorithm~\ref{alg:pseudo code}, which consists of the pre-training phase (Step 1-4) and the training phase (Step 6-11). The pre-training is employed to obtain a parameterized GNN. The pre-trained GNN then generates node representations $\mathbf{h}$, which are used to compute sample weights and corrected labels. After that, model parameters are updated according to the loss function: %Finally, after training converges, the network outputs predictions for each node. 
\begin{equation}
\label{objective function}
        \mathcal{J}_f = (1 - \alpha) \mathcal{J}_r + \alpha \mathcal{J}_c + \beta \mathcal{J}_p.
\end{equation}
Compared with GNNs with the standard cross entropy loss, the training of UnionNET incurs an extra computational complexity of $\mathcal{O}(|L|ml)$ to estimate node-level class distributions, where $|L|$ is number of labeled nodes, $m$ is number of classes, and $l$ is number of nodes including context nodes in the support set. %More details are provided in Appendix.

%The is mainly due to the calculation of structural loss $L_s$, involving matrix multiplication between anchor node representations and their context nodes and negative samples. The extra complexity is $\mathcal{O}(nml)$, where $n$ is number of nodes, $m$ is number of classes, and $l$ is number of nodes including context nodes and negative samples. This is linear with the number of graph nodes, which is reasonably acceptable.

\begin{algorithm}[th]
  \caption{Robust training for GNNs against label noise}
  \label{alg:pseudo code}
  \KwIn{Graph $G=\{\mathcal{V}, \mathcal{E}, \mathbf{X}\}$, node sets ${\mathcal{L}, \mathcal{U}}$, $\alpha$, $\beta$}
  \KwOut{label predictions}
  Initialize network parameters\; 
  
  \For{$t=0; t < epoches; t=t+1$}
  {
  \If{$t < $ start\_epoch}
  {
  pre-train the network according to  Eq.(\ref{eq:standardCE}); %$\mathcal{J}_{pre} = \mathcal{J}(f(\mathbf{h}_{\mathbf{x}}), \mathbf{y})$\;
  }
  \Else
  {
  Generate node representations $\mathbf{h}_{\mathbf{x}}$\;
  Construct support set S for each node $\hat{\mathbf{x}} \in \mathcal{L}$\;
  Aggregate labels to produce node-level class distribution $P(\mathbf{y}| \hat{\mathbf{x}}, S)$\;
  Compute weight $p_r(\hat{\mathbf{y}}|\hat{\mathbf{x}}, S)$ using Eq.(\ref{reweight})\;
  Generate corrected label $\mathbf{y}^c$ and its weight $p_c(\mathbf{y}^c|\hat{\mathbf{x}}, S)$ using Eq.(\ref{correction})\;
  Update parameters by descending gradient of Eq.(\ref{objective function})
  
  %$L_f = (1 - \alpha) L_r + \alpha L_c + \beta L_s + \gamma (L_e + L_p)$
  }
    %\While{\textcolor{red}{not converge}}
    %\While{not converge}
  }
 %\EndFor
 \Return{Label predictions}
\end{algorithm}

\section{Experiments}
\label{sec:experments}

%Our proposed method is validated through comparisons with competitive baselines and analyses on benchmark datasets.

%\subsection{Datasets and Baselines}

\noindent\textbf{Datasets and Baselines.} Three benchmark datasets are used in our experiments: Cora, Citeseer, and Pubmed\footnote{\label{linqs}https://linqs.soe.ucsc.edu/data}.
%Cora\footnote{\label{linqs}https://linqs.soe.ucsc.edu/data}, Citeseer\textsuperscript{\ref{linqs}}, and Pubmed\textsuperscript{\ref{linqs}}. %where documents are treated as nodes with bag-of-words feature vector, and citation links among documents are treated as undirected edges. 
We use the same data split as in~\cite{kipf2017semi}, with 500 nodes for validation, 1000 nodes for testing, and the remaining for training. Of these training sets, only a small fraction of nodes are labeled (3.6\% on Citeseer, 5.2\% on Cora, 0.3\% on Pubmed) and the rest of nodes are unlabeled. 
%each of which has the sparse feature vectors of bag-of-words. The citation links among these documents are treated as undirected edges. 
%The datasets are summarized in Table~\ref{dataset}. 
Details about the datasets can be found in \cite{kipf2017semi}.

% \begin{table}[h]
%     \centering
%     \label{dataset}
%     \caption{Details of datasets}
%     \scalebox{0.85}{
%     \begin{tabular}{cccccc}
%         \toprule
%         Dataset & Nodes & Edges & Classes & Features & Label rate\\
%         \hline
%         Citeseer & 3327 & 4732 & 6 & 3703 & 3.6\% \\
%         Cora & 2708 & 5429 & 7 & 1433 & 5.2\%\\
%         Pubmed & 19717 & 44338 & 3 & 500 & 0.3\%\\
%         %Dblp & 18447 & 91052 & 4 & 2476 & 0.4\%\\
%         \bottomrule
%     \end{tabular}}
%     \label{dataset}
% \end{table}

%\subsection{Baselines}
As far as we are concerned, there has not yet been any method exclusively proposed to deal with the label noise problem on GNNs for semi-supervised node classification. We select three strong competing methods from image classification, and adapt them to work with GCN~\cite{kipf2017semi} under our setting as baselines. 
\begin{itemize}\vspace{-4mm}
    \item \textbf{Co-teaching}~\cite{han2018co} trains two peer networks and each network selects the samples with small losses to update the other network. 
    \item \textbf{Decoupling}~\cite{malach2017decoupling} also trains two networks, but updates model parameters using only the samples with which two networks disagree.  
    \item \textbf{GCE}~\cite{zhang2018generalized} utilizes a negative Box-Cox transformation as the loss function. 
\end{itemize}\vspace{-4mm}

As a general robust training framework, UnionNET can be applied to any semi-supervised GNNs for node classification. Hereby, we instantiate UnionNET with two state-of-the-art GNNs, GCN~\cite{kipf2017semi} and GAT~\cite{velivckovic2017graph}, denoted as \textbf{UnionNET-GCN} and \textbf{UnionNET-GAT}, respectively.

%The three methods are as follows:
% \begin{itemize}
%     \item \textbf{Co-teaching}~\cite{han2018co} trains two peer networks and each network selects the samples with small losses to update the other network.
%     \item \textbf{Decoupling}~\cite{malach2017decoupling} also trains two networks, but updates model parameters using only the samples with which two networks disagree.
%     \item \textbf{GCE}~\cite{zhang2018generalized} utilizes a negative Box-Cox transformation as the loss function. %to improve the robustness of the standard cross entropy.
%     %as the loss function to relax the excessive attention that the standard cross entropy pays on the difficult samples. 
% \end{itemize}
 %Besides, the original GCN and GAT with standard cross entropy loss function are also used for comparison, which are denoted as \textbf{GCN-SCE} and \textbf{GAT-SCE}. 

%\subsection{Experimental Setup}
\noindent{\textbf{Experimental Setup.}} Due to the fact that there are not yet benchmark graph datasets corrupted with noisy labels, we manually generate noisy labels on public datasets to evaluate our algorithm. We follow commonly used label noise generation methods in the domain of images~\cite{han2018co,jiang2017mentornet}.
Given a noise rate $r$, we generate noisy labels over all classes according to a noise transition matrix $Q^{m\times m}$, where $Q_{ij} = p(\Tilde{y}=j|y=i)$ is the probability of clean label $\mathbf{y}$ being flipped to noisy label $\Tilde{\mathbf{y}}$. We consider two types of noise: 1). \textbf{Symmetric noise}: label $i$ is corrupted to other labels with a uniform random probability, s.t. $Q_{ij} = Q_{ji}$; 2). \textbf{Pairflip noise}: mislabeling only occurs between similar classes. For instance, given $r = 0.4$ and $m = 3$, the two types of noise transition matrices are given by %Eq.(\ref{symmetric}) and Eq.(\ref{pairflip}):
\begin{equation}
\label{symmetric}
    Q^{\text{symmetric}} = 
    \begin{bmatrix}
    0.6 & 0.2 & 0.2 \\
    0.2 & 0.6 & 0.2 \\
    0.2 & 0.2 & 0.6 
    \end{bmatrix}; \ \ 
    Q^{\text{pairflip}} = 
    \begin{bmatrix}
    0.6 & 0.4 & 0. \\
    0. & 0.6 & 0.4 \\
    0.4 & 0. & 0.6 \nonumber
    \end{bmatrix}
\end{equation}

% \begin{itemize}
% \item\textbf{Symmetric noise}: label $i$ is corrupted to other labels with a uniform random probability, s.t. $Q_{ij} = Q_{ji}$; 
% \item\textbf{Pairflip noise}: mislabeling only occurs between similar classes. 
% \end{itemize}

%\begin{equation}
%\label{pairflip}
%    Q^{\text{pairflip}} = 
%    \begin{bmatrix}
%    0.6 & 0.4 & 0. \\
%    0. & 0.6 & 0.4 \\
%    0.4 & 0. & 0.6 
%    \end{bmatrix}
%\end{equation}
Our experiments follow a transductive setting, where the noise transition matrix is only applied to $\mathcal{L}$, while both validation and test sets are kept clean. 
%, where all node features are available during training, whereas only nodes in $\mathcal{L}$ are used for training
%All models are trained on the corrupted validation data and tested on uncorrupted test data, with respect to two types of label noise.}
%We repeat all experiments are repeated for 10 times and report their mean values
For UnionNET-GCN, we apply a two-layer GCN, which has 16 units of hidden layer. The hyper-parameters are set as L2 regularization of $5*10^{-4}$, learning rate of 0.01, dropout rate of 0.5. For UnionNET-GAT, we apply a two-layer GAT, with the first layer consisting of 8 attention heads, each computing 8 features. The learning rate is 0.005, dropout rate is 0.6, L2 regularization is $5*10^{-4}$.

We set the random walk length as 10 on Cora and Citeseer, and 4 on Pubmed, and the random walk is repeated for 10 times for each node to create the support set. We first pre-train the network, during which only the standard cross entropy are used, i.e. $\mathcal{J}_{pre} = \mathcal{J}(f(\mathbf{h}_{\mathbf{x}}), \mathbf{y})$. After that, it proceeds to the formal training, which uses $\mathcal{J}_f$ in Eq.(\ref{objective function}) as the loss function. And $\alpha$ and $\beta$ are set as 0.5 and 1.0. 

\subsection{Comparison with State-of-the-art Methods}
\begin{table*}[t]
    \centering
    \footnotesize 
    \tabcolsep 10pt
    \caption{Performance comparison (Micro-F1 score) on node classification}
    \scalebox{0.7}{
    \begin{tabular}[c]{c|c|cccc|cccc}
         \toprule
         & & \multicolumn{4}{c}{Symmetric label noise} & \multicolumn{4}{c}{Asymmetric label noise} \\
         \cline{3-10}
         Dataset & Methods & \multicolumn{8}{c}{noise rate (\%)} \\
         \cline{3-10}
          &  & 10 & 20 & 40 & 60 & 10 & 20 & 30 & 40 \\
         \hline
         \multirow{7}{1cm}{Cora}&
         GCN &0.778 &0.732 & 0.576 & 0.420 &0.768 & 0.696&0.636& 0.517 \\
         &Co-teaching &0.775 &0.665 & 0.486 & 0.249 & 0.773 & 0.630 & 0.542 & 0.393\\
         &Decoupling &0.738 &0.708 & 0.564 & 0.436 & 0.743 & 0.683 & 0.574 & 0.518\\
         &GCE &{0.794} &0.741 & 0.621 & 0.402 & 0.773 & 0.714 & 0.652 & 0.509\\
         &UnionNET-GCN &\textbf{0.812} &\textbf{0.795} & \textbf{0.707} & \textbf{0.491} & \textbf{0.801} & \textbf{0.771} & \textbf{0.710} & \textbf{0.584}\\
         \cline{2-10}
         &GAT & 0.755 &0.709 & 0.566 & 0.389 &0.764 & 0.683 &0.616 & 0.534 \\
         &UnionNET-GAT & {0.797} &0.784 & 0.692 & 0.546 &0.774 & 0.745 &0.660 & 0.540 \\
         \hline
         \hline
         \multirow{7}{1cm}{Citeseer}&
         GCN &0.670 &0.634 & 0.480 & 0.360 &0.667 & 0.624 & 0.531 & 0.501 \\
         &Co-teaching &0.673 &0.541 & 0.379 & 0.273 & 0.677 & 0.583 & 0.472 & 0.418\\
         &Decoupling &0.588 &0.584 & 0.402 & 0.348 & 0.615 & 0.548 & 0.537 & 0.468\\
         &GCE &0.690 &0.649 & 0.542 & 0.358 & {0.701} & 0.633 & 0.552 & 0.498\\
         &UnionNET-GCN &\textbf{0.701} &\textbf{0.673} & \textbf{0.567} & \textbf{0.401} & \textbf{0.706} & \textbf{0.667} & \textbf{0.587} & \textbf{0.521}\\
         \cline{2-10}
         &GAT & 0.649 &0.604 & 0.475 & 0.338 &0.651 & 0.599 &0.551 & 0.480 \\
         &UnionNET-GAT & {0.695} &0.667 & 0.585 & 0.424 &0.697 & 0.654 &0.604 & 0.512 \\
         \hline\hline
         \multirow{7}{1cm}{Pubmed}&
         GCN &0.748 &0.672 & 0.508 & 0.367 &0.739 & 0.686 & 0.618 & 0.528 \\
         &Co-teaching &0.769 &0.660 & 0.478 & 0.345 & 0.761 & 0.634 & 0.576 & 0.472\\
         &Decoupling &0.650 &0.625 & 0.422 & 0.334 & 0.641 & 0.592 & 0.428 & 0.396\\
         &GCE &0.750 &0.699 & 0.561 & 0.393 & 0.753 & 0.696 & 0.609 & {0.567}\\
         &UnionNET-GCN &\textbf{0.769} &\textbf{0.725} & \textbf{0.588} & \textbf{0.409} & \textbf{0.776} & \textbf{0.719} & \textbf{0.649} & \textbf{0.556}\\
         \cline{2-10}
         &GAT & 0.736 &0.670 & 0.525 & 0.381 &0.737 & 0.657 &0.594 & 0.536 \\
         &UnionNET-GAT & {0.751} &0.726 & 0.570 & 0.361 &0.758 & 0.702 &0.626 & 0.552 \\
         \bottomrule
    \end{tabular}}\vspace{-4mm}
    \label{table:comparison}
\end{table*}

Table~\ref{table:comparison} compares the node classification performance of all methods w.r.t. both the symmetric and asymmetric noise types under various noise rates. The best performer is highlighted by \textbf{bold} on each setting. 
%The hyper-parameters are set as follows: 1) for UnionNET-GCN, $\alpha=0.5$, $\beta=0.2$, $\gamma=0.5$, except that $\beta$ is set to 0.5 on Pubmed in response to its higher structural clustering property; 2) for UnionNET-GAT, $\alpha=0.5$, $\beta=0.5$, $\gamma=0.5$. All experiments are repeated for 10 times, and the mean values of their Micro-F1 scores are reported here. 
For GCN-based baselines, UnionNET-GCN generally outperforms all baselines by a large margin. Compared with GCN in case of symmetric noise type, UnionNET-GCN achieves an accuracy improvement of 3.4\%, 6.3\%, 13.1\% and 7.1\% under the noise rate of 10\%, 20\%, 40\% and 60\% on Cora, respectively. Similar improvements can be seen on Citeseer and Pubmed, where the smallest improvement is 2.1\% on Pubmed with a noise rate of 10\%, and the largest improvement is 8.7\% on Citeseer with a noise rate of 40\%. In case of asymmetric noise type, UnionNET-GCN has the similar performance. Quantitatively, UnionNET-GCN outperforms GCN by an average of 3.6\%, 5.0\%, 5.4\%, 3.8\% on the four noise rates on three datasets.  %Union-GCN, although not like in the symmetric noise case where it exhibits 
%significant superiority on high noise rates, it still tends to achieve better performance on higher noise rates.

In most cases, GCE is the second best performer, %because its new loss function relaxes the sensitivity of the cross entropy with respect to corrupted labels. 
but its advantage comes at the cost of worse converging capability, leading to sub-optimal performance. Co-teaching and Decoupling do not exhibit robustness towards noisy labels as reported in fully supervised image classification. Their performance drops are expected, as labeled data is further reduced when they prune the training data. This exacerbates the label scarcity problem in our semi-supervised setting.

%This is because, they selectively discards a quantity of samples with relatively larger training loss, while Decoupling only keeps samples with which the two networks disagree.

%4\% on Cora, 5.7\% on Citeseer, and 5.3\% on Pubmed over the four noise rates. 
%When they prune the training set, they discard much useful information at the same time. More importantly, they reduce the number of accessible labels, which exacerbates the label scarcity problem in our semi-supervised setting.

On three datasets, UnionNET-GAT also surpasses GAT w.r.t. most noise rates. Similar to UnionNET-GCN, UnionNET-GAT generally exhibits greater superiority on higher noise rates. For example, in case of symmetric noise type, UnionNET-GAT outperforms GAT by an average of 3.4\%, 6.4\%, 9.4\% and 7.4\% at the four noise rates. 
%And it achieves the largest improvement of 15.7\% on Cora at a 60\% noise rate. 
Such performance gains validate the generality of UnionNET on improving robustness of different GNN models against noisy labels.

%In case of asymmetric noise type, UnionNET-GAT achieves average improvements of 2.4\%, 6.0\%, 6.9\%, 4.1\% at the four noise rates.

%Although GCE applies a new loss function that relaxes the sensitivity of the cross entropy w.r.t. incorrect labels, it sacrifices its converging capability, thereby leading to suboptimal performance.

%In summary, UnionNET yields superior robustness to both symmetric and asymmetric noise types over competing methods and offers remarkable improvement gains on higher noise rates.

%\begin{figure*}[!htbp]\label{figure:sensitivity test}
%    \centering
%    \subfigure[Cora]{
%        \centering
%        \includegraphics[width=5.7cm]{pics/parameter_sensitivity_cora_sym.png}}
%        \subfigure[Citeseer]{
%        \centering
%        \includegraphics[width=5.7cm]{pics/parameter_sensitivity_citeseer_sym.png}}
%        \subfigure[Pubmed]{
%        \centering
%        \includegraphics[width=5.7cm]{pics/parameter_sensitivity_pubmed_sym.png}}
%        \vspace{-0.2cm}
%    \small{\caption{Hyper-parameters sensitivity test on Cora with 40\% symmetric noise rate}}
%\end{figure*}

\subsection{Ablation Study}

We conduct ablation studies to test the effectiveness of different components in UnionNET. Our ablation study is based on GCN, with two ablation versions: 1) \textbf{UnionNET-R} with only sample reweighting; 2) \textbf{UnionNET-RC} with sample reweighting and label correction. The ablation results are summarized in Table~\ref{table:ablation}. When only reweighting is applied, UnionNET-R consistently exhibits advantages over GCN, though the advantageous margins vary over different noise rates and noise types. %Averagely, UnionNET-R surpasses 3.0\%, 0.94\% and 3.5\% over GCN on the three datasets.
%, where it achieves the highest superiority of 13.5\% on Cora at the 60\% of symmetric noises. 
When it comes to UnionNET-RC, both smaple reweighting and label correction are applied, but, surprisingly, the performance becomes worse than UnionNET-R in some extreme cases with higher noise rates. Therefore, label correction does not guarantee performance gains, whose utility is exerted only with the regularization of the prior distribution loss.

\label{ablation study}
\begin{center}
    \begin{table*}[t]
    \centering
    \footnotesize 
    \tabcolsep 6pt
    \caption{Performance comparison of ablation experiments based on GCN}
        \scalebox{0.7}{
    \begin{tabular}[c]{c|c|cccc|cccc}
        \toprule
        & & \multicolumn{4}{c}{Symmetric label noise} & \multicolumn{4}{c}{Asymmetric label noise} \\
        \cline{3-10}
        Dataset & Methods & \multicolumn{8}{c}{noise rate (\%)} \\
        \cline{3-10}
         & &10 & 20 & 40 & 60 & 10 & 20 & 30 & 40 \\
        \hline
        \multirow{5}{1cm}{Cora}&
        GCN &0.778 &0.732 & 0.576 & 0.420 &0.768 & 0.696&0.636& 0.517 \\
        &UnionNET-R &0.785 &{0.770} & 0.659 & 0.480 & {0.796} & 0.709 & 0.646 & 0.521\\
        &UnionNET-RC &{0.788} &0.759 & 0.626 & 0.339 & 0.783 & 0.703 & 0.601 & 0.516\\
        &UnionNET-GCN &\textbf{0.812} &\textbf{0.795} & \textbf{0.707} & \textbf{0.491} & \textbf{0.801} & \textbf{0.771} & \textbf{0.710} & \textbf{0.584}\\
        \hline
        \hline
        \multirow{5}{1cm}{Citeseer}&
        GCN &0.670 &0.634 & 0.480 & 0.360 &0.667 & 0.624 & 0.531 & 0.501 \\
        &UnionNET-R &0.692 &0.643 & 0.507 & 0.363 & 0.699 & 0.627 & 0.547 & 0.484\\
        &UnionNET-RC &0.657 &0.645 & 0.495 & 0.330 & 0.660 & 0.642 & 0.511 & 0.431\\
        &UnionNET-GCN &\textbf{0.701} &\textbf{0.673} & \textbf{0.567} & \textbf{0.401} & \textbf{0.706} & \textbf{0.667} & \textbf{0.587} & \textbf{0.521}\\

        \hline
        \hline
        \multirow{5}{1cm}{Pubmed}&
        GCN &0.748 &0.672 & 0.508 & 0.367 &0.739 & 0.686 & 0.618 & 0.528 \\
        &UnionNET-R &0.766 &0.710 & 0.573 & \textbf{0.417} & 0.759 & 0.705 & 0.624 & 0.560\\
        &UnionNET-RC &\textbf{0.770} &{0.695} & 0.573 & 0.362 & 0.757 & 0.650 & 0.608 & {0.497} \\
        &UnionNET-GCN &{0.769} &\textbf{0.725} & \textbf{0.588} & {0.409} & \textbf{0.776} & \textbf{0.719} & \textbf{0.649} & \textbf{0.556}\\

        \bottomrule
    \end{tabular}}\vspace{-4mm}
    \label{table:ablation}
\end{table*}
\end{center}

\vspace{-1.3cm}
\subsection{Hyper-parameter Sensitivity}
\begin{figure*}[!htbp]
    \centering\vspace{-4mm}
    \subfigure[$\alpha$]{
        \centering
        \includegraphics[width=3.8cm]{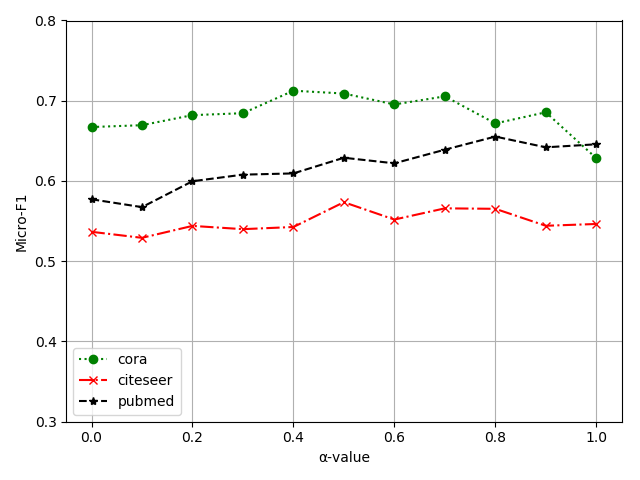}}
        \subfigure[$\beta$]{
        \centering
        \label{figure:sensitivity-beta}
        \includegraphics[width=3.8cm]{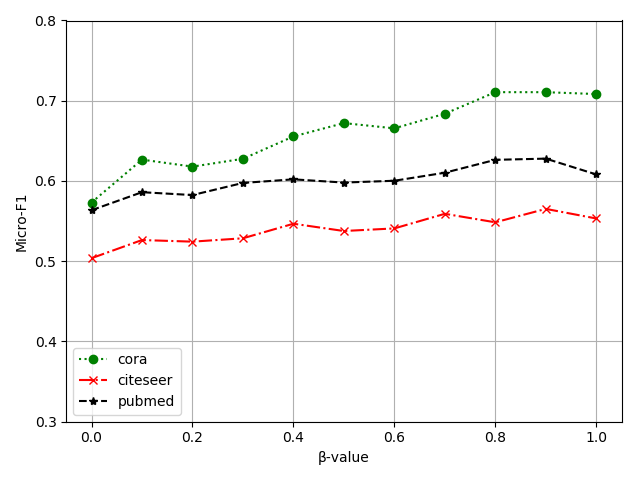}}
        \subfigure[Walk Length]{
        \centering
        \includegraphics[width=3.8cm]{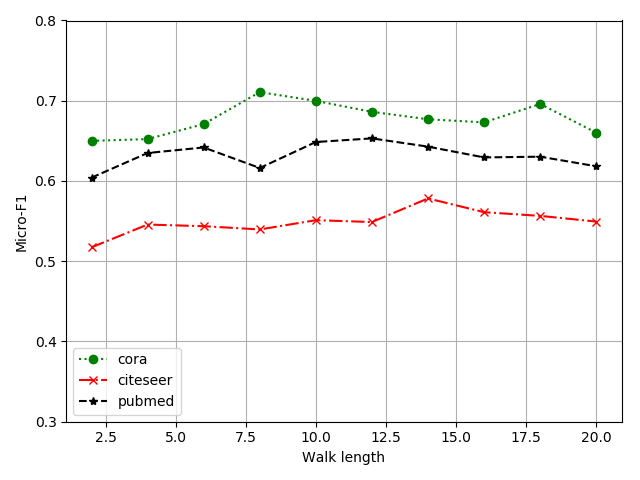}}
        \vspace{-0.3cm}
    \caption{Hyper-parameter sensitivity analysis on $\alpha, \beta$, and the random walk length}
    \label{figure:sensitivity}
\end{figure*}

We further test the sensitivity of UnionNET-GCN w.r.t. the hyper-parameters ($\alpha$, $\beta$) in Eq.(\ref{objective function}) and the random walk length for the support set construction. We report the results on the three datasets at 40\% symmetric noise rate in Fig.~\ref{figure:sensitivity}. $\alpha$ controls the trade-off between sample reweighting and label correction. When $\alpha$ is zero, our method is only a reweighting method. When $\alpha$ reaches $1$, our method evolves as a self-learning based label correction method, where given labels are replaced with predicted labels after the initial epoches. On Cora and Citeseer, our method achieves the best results at a medium $\alpha$ value. But on Pubmed, its performance improves as $\alpha$ increases, and reaches its best when $\alpha = 1.0$. This is possibly because Pubmed has stronger clustering property with only three classes, enabling the predicted labels to be more reliable for correction. The performance changes w.r.t. $\beta$ exhibits similar trends on the three datasets, where our method gradually improves its performance as $\beta$ increases. The random walk length determines the order of proximity the support set could cover. Either too small or too large of the random walk length would impair the reliability of the supportive nodes, and thus undermine performance improvements. Empirically, our method achieves its best at a medium range of random walk lengths.

% The similar trends w.r.t. $\beta$ and $\gamma$ can be observed on Citeseer. $\beta$ controls how much contribution the structural loss $L_s$ makes to our method, while $\gamma$ takes the responsibility of regularization during label correction. When their values are very small, they are too weak to diminish the negative impact of noisy labels and the dual label supervision. As they increase, the network starts to achieve better performance and stays stabilized. 

\vspace{-1cm}
\section{Conclusion}
We proposed a novel semi-supervised framework, UnionNET, for learning with noisy labels on graphs. We argued that, existing methods on image classification fail to work on graphs, as they often take a fully supervised approach, and requires extra clean supervision or explicit estimation of the noise transition matrix. Our approach provides a unified solution to robustly training a GNN model and performing label correction simultaneously. %The key idea is to perform label aggregation to estimate node-level class probability distributions, which are used to guide sample reweighting and label correction.
UnionNET is a general framework that can be instantiated with any state-of-the-art semi-supervised GNNs to improve model robustness, and it can be trained in an end-to-end manner. Experiments on three real-world datasets demonstrated that our method is effective in improving model robustness w.r.t. different label noise types and rates, and outperform competitive baselines.

\vspace{1mm}
%\subsection*{Acknowledgement}
\noindent{\textbf{Acknowledgement.}}
This work is supported by the USYD-Data61 Collaborative Research Project grant, the Australian Research Council under Grant DP180100966, and the China Scholarship Council under Grant 201806070131.
%\small
\bibliographystyle{splncs04}
\bibliography{reference.bib}

\begin{thebibliography}{10}
\providecommand{\url}[1]{\texttt{#1}}
\providecommand{\urlprefix}{URL }
\providecommand{\doi}[1]{https://doi.org/#1}

\bibitem{de2020analysis}
de~Aquino~Afonso, B.K., Berton, L.: Analysis of label noise in graph-based
  semi-supervised learning. In: SAC. pp. 1127--1134 (2020)

\bibitem{arazo2019unsupervised}
Arazo, E., Ortego, D., Albert, P., O'Connor, N.E., McGuinness, K.: Unsupervised
  label noise modeling and loss correction. In: ICML. pp. 312--321 (2019)

\bibitem{gilmer2017neural}
Gilmer, J., Schoenholz, S.S., Riley, P.F., Vinyals, O., Dahl, G.E.: Neural
  message passing for quantum chemistry. In: ICML. pp. 1263--1272 (2017)

\bibitem{goldberger2016training}
Goldberger, J., Ben-Reuven, E.: Training deep neural-networks using a noise
  adaptation layer. In: ICLR (2016)

\bibitem{hamilton2017inductive}
Hamilton, W., Ying, Z., Leskovec, J.: Inductive representation learning on
  large graphs. In: NeurIPS. pp. 1024--1034 (2017)

\bibitem{han2018co}
Han, B., Yao, Q., Yu, X., Niu, G., Xu, M., Hu, W., Tsang, I., Sugiyama, M.:
  Co-teaching: Robust training of deep neural networks with extremely noisy
  labels. In: NeurIPS. pp. 8527--8537 (2018)

\bibitem{han2019deep}
Han, J., Luo, P., Wang, X.: Deep self-learning from noisy labels. In: ICCV. pp.
  5138--5147 (2019)

\bibitem{jiang2017mentornet}
Jiang, L., Zhou, Z., Leung, T., Li, L.J., Fei-Fei, L.: {M}entor{N}et: Learning
  data-driven curriculum for very deep neural networks on corrupted labels. In:
  ICML. pp. 2304--2313 (2018)

\bibitem{karimi2020deep}
Karimi, D., Dou, H., Warfield, S.K., Gholipour, A.: Deep learning with noisy
  labels: Exploring techniques and remedies in medical image analysis. Medical
  Image Analysis  \textbf{65},  101759 (2020)

\bibitem{kipf2017semi}
Kipf, T.N., Welling, M.: Semi-supervised classification with graph
  convolutional networks. In: ICLR (2017)

\bibitem{li2018deeper}
Li, Q., Han, Z., Wu, X.M.: Deeper insights into graph convolutional networks
  for semi-supervised learning. In: AAAI. pp. 3538--3545 (2018)

\bibitem{lu2003link}
Lu, Q., Getoor, L.: Link-based classification. In: ICML. pp. 496--503 (2003)

\bibitem{malach2017decoupling}
Malach, E., Shalev-Shwartz, S.: Decoupling ``when to update" from ``how to
  update". In: NeurIPS. pp. 960--970 (2017)

\bibitem{nt2019learning}
NT, H., Jin, C., Murata, T.: Learning graph neural networks with noisy labels.
  In: 2nd ICLR Learning from Limited Labeled Data (LLD) Workshop (2019)

\bibitem{patrini2017making}
Patrini, G., Rozza, A., Krishna~Menon, A., Nock, R., Qu, L.: Making deep neural
  networks robust to label noise: A loss correction approach. In: CVPR. pp.
  1944--1952 (2017)

\bibitem{ren2018learning}
Ren, M., Zeng, W., Yang, B., Urtasun, R.: Learning to reweight examples for
  robust deep learning. In: ICML. pp. 4334--4343 (2018)

\bibitem{sukhbaatar2014training}
Sukhbaatar, S., Bruna, J., Paluri, M., Bourdev, L., Fergus, R.: Training
  convolutional networks with noisy labels. In: ICLR (2014)

\bibitem{tanaka2018joint}
Tanaka, D., Ikami, D., Yamasaki, T., Aizawa, K.: Joint optimization framework
  for learning with noisy labels. In: CVPR. pp. 5552--5560 (2018)

\bibitem{vahdat2017toward}
Vahdat, A.: Toward robustness against label noise in training deep
  discriminative neural networks. In: NeurIPS. pp. 5596--5605 (2017)

\bibitem{velivckovic2017graph}
Veli{\v{c}}kovi{\'c}, P., Cucurull, G., Casanova, A., Romero, A., Lio, P.,
  Bengio, Y.: Graph attention networks. In: ICLR (2018)

\bibitem{zhang2017mixup}
Zhang, H., Cisse, M., Dauphin, Y.N., Lopez-Paz, D.: mixup: Beyond empirical
  risk minimization. ICLR  (2018)

\bibitem{zhang2018generalized}
Zhang, Z., Sabuncu, M.: Generalized cross entropy loss for training deep neural
  networks with noisy labels. In: NeurIPS. pp. 8778--8788 (2018)

\bibitem{zhu2003semi}
Zhu, X., Ghahramani, Z., Lafferty, J.D.: Semi-supervised learning using
  gaussian fields and harmonic functions. In: ICML. pp. 912--919 (2003)

\end{thebibliography}
\end{document}